# Sparsifying Spiking Networks through Local Rhythms

Wilkie Olin-Ammentorp
Mathematics and Computer Science Division
Argonne National Laboratory
Lemont, IL, USA
wolinammentorp@anl.gov

## ABSTRACT

It has been well-established that within conventional neural networks, many of the values produced at each layer are zero. In this work, I demonstrate that spiking neural networks can prevent the transmission of spikes representing values close to zero using local information. This can reduce the amount of energy required for communication and computation in these networks while preserving accuracy. Additionally, this demonstrates a novel application of biologically observed spiking rhythms.

## CCS CONCEPTS

• Computing methodologies – Machine Learning – Machine learning approaches – Bio-inspired approaches

• Computing methodologies – Artificial Intelligence

## KEYWORDS

Spiking neural networks, resonate and fire neurons, analog neural networks, vector symbolic architectures, sparsity, neuromorphic computing, high dimensional computing

## 1 Introduction

It has been experimentally established that deep neural networks (including multi-layer perceptrons, convolutional networks, transformers, and more) exhibit sparsity in both their weights and activation values. This has led to challenges as well as opportunities, and programmers and computer architects continue to consider techniques to take advantage of this sparsity [1]–[4].

Similarly, the mammalian brain exhibits many forms of sparsity. Neurons in the brain follow a log-normal distribution of firing rates; most neurons fire less than once every second, and approximately half of all neurons fire even less often [5]. This raises the question of how the brain employs sparsity in its operation, and whether any principles derived from its operation can be applied to allow artificial neural networks to operate more efficiently.

I propose that applying phase encoding to construct spiking neural networks naturally yields a sparsification method by assuming that any missing values are in-phase with a reference oscillation. Omitting these values allows approximately equivalent inferences to be carried out while requiring less communication.

### 1.1 Phase Coding

Much of the information in the brain is represented using electrical impulses generated by neurons which are commonly called 'spikes.' Spikes can be used to encode information in a number of different ways, from 'rate coding,' in which the number of spikes within a given period encodes a value, to 'inter-spike intervals,' in which the timing between two spikes is used to encode a value [6].

'Phase coding' uses spikes to represent angular values. These angles represent how far in or out of phase a neuron's internal potential oscillates with respect to a reference signal (Figure 1). Furthermore, these angles can be represented explicitly with floating points or with precisely timed spikes. In the latter case, spikes represent when a neuron's potential reaches a local maximum.

Within the brain, many oscillations on varying spatial and temporal domains are present to serve as reference signals against which the phase of a neuron can be compared [7]. These systemic oscillations form a crucial aspect of computation in the brain, emerging spontaneously in cortical tissue and providing functionality which vital to computation, rather than an emerging as a side-effect [8]–[10].

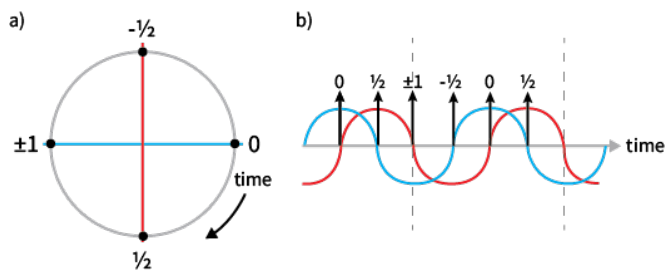

**Figure 1:** An illustration of phase encoding. Spikes represent positions around the unit circle in the complex plane (a), which can be represented by an angle or a complex value. The value of a spike can be determined by comparing to a real or complex valued oscillation (b).

In this work, I apply phase coding to execute artificial neural networks via spikes. This technique has been previously demonstrated, allowing for neural networks which can be trained conventionally in the complex domain and executed via networks of spiking resonate-and-fire neurons [11], [12]. Furthermore, the correspondence of phase-based representations with what is known as a 'vector-symbolic system' allows vectors of activations to be combined and manipulated algebraically. This ability to describe, combine, and manipulate phase-based information at a high level enables the construction of deep, complex networks which operate similarly to residual and attention-based architectures [13]. The utilization of phase-based information at all layers in these networks yields a flexible system which can compute by exchanging packets of precisely-timed binary spikes. This encoding allows for values to be defined on a short timeframe, as well as the ability to average across multiple packets, and it mimics the transmission of information as observed in the brain [14], [15].

Experimentally, this representation has been able to scale to large tasks and offers a trade-off between the inefficiency exhibited by previous rate-coding approaches and the fragility of time-coding approaches [11]. However, sparsification of phasor networks offers a further path towards increasing efficiency by reducing communication overheads.

### 1.2 Sparsifying Inputs via Local Oscillations

In my previous work, an additional 'bias' input was applied to phasor neurons [13]. This change shifted the distribution of activations – each representing an angle - from uniform across the unit circle to normally-distributed around zero (Figure 2a). This change allowed for the creation of residual blocks of phasor neurons – layers which approximate the identity function when initialized. This enabled very deep phasor networks which were successfully trained with backpropagation [3], [13].

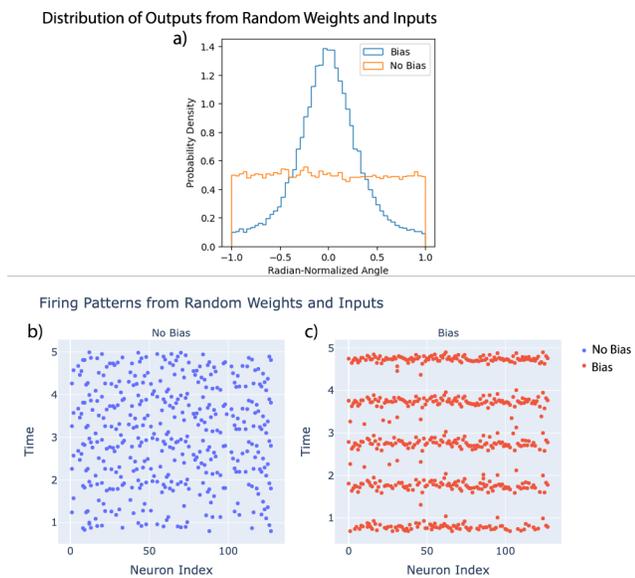

**Figure 2:** Firing patterns produced by a phase-based spiking neural network layer with random weights and inputs. A layer without a bias produces spikes with uniformly distributed phase, but when a phase bias is introduced, spikes become normally around a reference phase.

This change also led to different behavior in the spiking domain. When angles are viewed as spiking outputs, the complex representation of each phase is 'unrolled' into the temporal domain. In this case, the shift of distributions from uniform (Figure 2b) to normal (Figure 2c) can be seen in the 'bunching' of spikes, as firing becomes significantly more likely at certain times than others.

With so many spikes being produced at the same time in a layer with biased neurons, I propose that neurons which spike out-of-phase produce information which is more valuable to a downstream input. Spikes which are approximately in-phase with the larger rhythm can be eliminated, and downstream neurons can assume this 'missing' input simply represents the previous group's average phase.

## 2 Results

### 2.1 Experimental Setup

I investigate the hypothesis that most in-phase spikes are redundant by eliminating spikes from the execution of a phasor neural network. This network is a multi-layer perceptron executing the FashionMNIST image classification task [16]. The 768 input pixels are first randomly projected into 1024 values in the phasor domain using non-trainable random weights. This is followed by two layers of 128 and 10 phasor neurons, each with a bias input. I examine the performance of this network as it undergoes different methods of sparsification: one method removes spikes on their reconstructed phase, another removes spikes based on a local oscillation, and one removes spikes at random.

My current implementation of phasor networks allows for two inference modes: one carried out by the exchange of floating-point tensors which explicitly represent phase values, and one carried out by exchanging lists of times representing binary spikes ("spike trains"). Networks are trained in the non-spiking mode using standard backpropagation through phasor neurons. The loss function at the output layer is the cosine similarity between network's output and the image's label represented by a phase-shift keying of 90°. The RMSProp optimizer (learning rate 0.001) is used to decrease the loss function through 1000 batches of 128 images from the training set.

In the spiking inference mode, each phasor neuron is executed identically to a resonate-and-fire (R&F) neuron model [17]. The phase-encoded inputs are converted to spikes and presented three times, stimulating the layers of phasor neurons. These neurons oscillate in the current-voltage domain, spiking when voltage reaches a maximum value and is above a certain threshold. The phase of output spikes is decoded at the final layer at the time of its $5^{th}$ spiking cycle, enabling the predicted label and accuracy to be calculated. Details on specific parameters such as neuron threshold, resonant frequency, and more are identical to my previous work [11].

### 2.2 Sparsification Methods

All sparsification methods are applied while the network is executing in its spiking inference mode. These methods are applied at the output of each phasor layer with trainable input weights. Each method takes the spike train produced by the phasor layer and selectively removes a subset based on different criteria.

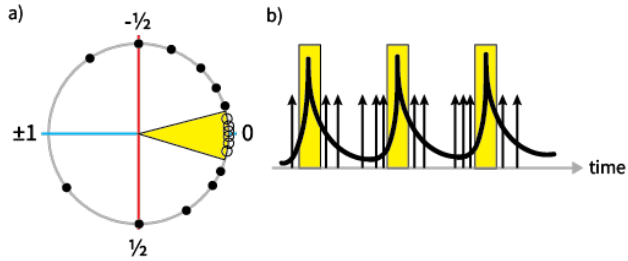

**Figure 3:** Two mechanisms for sparsifying activations in phasor neural networks by removing values in-phase with a reference oscillation. Values represented by angles (a) can simply be removed when within an arc of phase (yellow). Alternately, a local signal can be used to selectively inhibit spikes within a period of maximum activity (yellow, b).

#### 2.2.1 Explicit Sparsification

Phasor values lie on the unit circle in the complex plane, with values in-phase to a reference at 0, values out-of-phase at $\pm 1/2$, and anti-phase values at $\pm 1$ (Figure 1). In this method, spikes are explicitly decoded to phases using the network's global starting time and the neuron's depth in the network. Spikes which correspond to a phase value within an arc subtended from 0 are removed from the spike train (Figure 3a). The angle of the arc subtended determines the proportion of spikes removed from the train: an angle of 0 removes no spikes, and an angle of $2\pi$ removes all spikes.

#### 2.2.2 Inhibitory Sparsification

There is no absolute execution time or well-defined depth within a biological neural network which would allow it to execute the previous method. However, a multitude of other information is locally available to neurons in the oscillation of electric fields, individual ionic concentrations, neurotransmitters, and more. I emulate the use of a local field potential to inhibit a neuron's output. Each spike in a layer contributes to this field, and as the rate of change in this field reaches a local maximum, neurons in the layer are silenced. The duration of the inhibitory period centered on this maximum determines the proportion of spikes which are removed from the spike train (Figure 3b). An inhibitory period of 0 seconds removes no spikes, and an inhibitory period of 1.0 seconds (the full resonant period of the neuron) removes all spikes.

#### 2.2.3 Random Sparsification

A randomly chosen proportion of spikes is removed from the spike train to serve as a control method of sparsification.

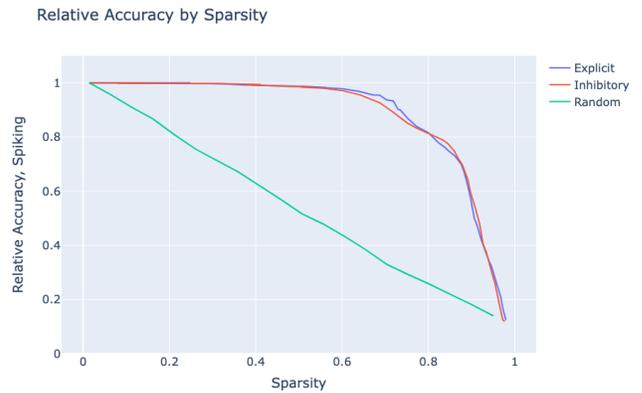

**Figure 4:** Performance of the MLP network on the FashionMNIST task as spike trains are sparsified by 3 methods. Methods which selectively remove spikes in-phase with the overall network rhythm greatly outperform a control method which randomly removes spikes.

### 2.3 Performance

As the parameter of each sparsification technique was varied, the performance of the network with respect to its initial accuracy (85.3%) was measured. The control method (random removal) displayed linear scaling of performance with the proportion of spikes removed. In contrast, both phase-based methods of sparsification displayed a markedly different trend: performance of the networks changed little until approximately 60% of spikes had been removed (Figure 4). This supports the hypothesis that many spikes are redundant when neurons in a layer are biased to spontaneously fire in-phase with one another.

## 3 Discussion

The sparsity of activations in neural networks provides an opportunity to reduce the burden of computation and communication required to execute them. However, realizing innately sparse computation remains a challenge. I suggest that phase-based spiking neural networks offer a potential pathway to meet this challenge. By organizing neurons into local groups (layers) which tend to fire in-phase with respect to a local oscillation, many of the spikes representing this information are redundant and can be removed while maintaining the correct overall network activity.

Currently, I achieve this behavior in an artificial manner: each neuron in a layer is 'biased' by a spike from a reference neuron at a particular time, essentially attaching each layer to a global clock. Additionally, spikes are removed from neurons 'post-hoc' to sparsify outputs, rather than being suppressed internally through neuronal dynamics. In future work, I aim to achieve these processes more organically by feedback connections within a layer to emulate the excitatory-inhibitory dynamics observed within biological neural networks [18].

Furthermore, the demonstration network is small and shallow. More work is needed to demonstrate that these findings will carry over to deeper networks with more complex architectures and tasks.

I aim to achieve this by extending my current implementation to execute phase-based residual and attentional networks fully in the spiking domain. This will allow further experiments to test whether phase-based sparsification methods will remain practical for more complex architectures and tasks.

## 4  Conclusion

The brain employs a rich variety of coding methods to transmit and manipulate information; each carries a set of advantages and disadvantages. Phase-coding is one of these techniques, and the paramount importance of rhythms within the brain suggests that it may be one of the most crucial to overall functionality.

In this work, I propose that an additional advantage of phase-coding is the ability to easily enable networks to operate with sparse inputs. Spikes which are in-phase with the network rhythm are mostly redundant, and out-of-phase spikes are rarer, suggesting it is more important to communicate the latter. Experimental evidence supports this, as removing in-phase spikes allowed for a significant reduction in the number of spikes transmitted through a network while maintaining performance.

In the future, I aim to demonstrate this principle holds for wider, deeper networks with more complex mechanisms, providing a pathway towards sparse and effective neural networks with links to biology and the potential for execution on hardware utilizing analog, oscillatory hardware.

## 5  Methods

All models were built using JAX 0.3.25 with Haiku 0.0.9. Code for models and experiments is publicly available at https://github.com/wilkieolin/phasor_jax.git

### ACKNOWLEDGMENTS

This work was supported by DOE ASCR and BES Microelectronics Threadwork. This material is based upon work supported by the U.S. Department of Energy, Office of Science, under contract number DE-AC02-06CH11357.

I would like to thank Angel Yanguas-Gil for his feedback on this work.